\documentclass{article}

\usepackage{times}
\usepackage{graphicx} 
\usepackage{subfigure} 
\usepackage{color}

\usepackage{natbib}

\usepackage{algorithm}
\usepackage{epstopdf}

\usepackage{hyperref}

\usepackage{url}

\usepackage{times}
\usepackage{epsfig}
\usepackage{amsmath}
\usepackage{amssymb}
\usepackage{tikz}
\usepackage{multirow}
\usepackage[noend]{algorithmic}
\usepackage{pbox}
\usepackage{enumitem}

\def\LT#1{{{#1}}}
\def\LTR#1{}

\def\DT#1{{{#1}}}
\def\DTR#1{}

\def\myparagraph#1{\noindent{\bf #1~~}}

\newcommand{\bx}{{\bf x}}
\newcommand{\bphi}{\mbox{\boldmath $\phi$}}
\newcommand{\bw}{{\bf w}}
\usepackage[accepted]{icml2014}

\icmltitlerunning{EXMOVES: Classifier-based Features for Scalable Action Recognition}
\begin{document}

\twocolumn[
\icmltitle{EXMOVES: Classifier-based Features for Scalable Action Recognition}

\icmlauthor{Du Tran, Lorenzo Torresani}{\{dutran,lorenzo\}@cs.dartmouth.edu}
\icmladdress{Computer Science Department,
            Dartmouth College, NH 03755 USA}


\vskip 0.3in
]

\begin{abstract} 
This paper introduces EXMOVES, learned exemplar-based features for efficient recognition of actions in videos. The entries in our descriptor are produced by evaluating a set of movement classifiers over spatial-temporal volumes of the input sequence. Each movement classifier is a simple exemplar-SVM trained on low-level features, i.e., an SVM learned using a single annotated positive space-time volume and a large number of unannotated videos.

Our representation offers two main advantages. First, since our mid-level features are learned from individual video exemplars, they require minimal amount of supervision. Second, we show that simple {\em linear} classification models trained on our global video descriptor yield action recognition accuracy approaching the state-of-the-art but at orders of magnitude lower cost, since at test-time no sliding window is necessary and linear models are efficient to train and test. This enables scalable action recognition, i.e., efficient classification of a large number of actions even in massive video databases. We show the generality of our approach by building our mid-level descriptors from two different low-level feature vectors. The accuracy and efficiency of the approach are demonstrated on several large-scale action recognition benchmarks.
\end{abstract} 

\section{Introduction}
Human action recognition is an important but still largely unsolved problem in computer vision with many potential useful applications, including content-based video retrieval, automatic surveillance, and human-computer interaction. The difficulty of the task stems from the large intra-class variations in terms of subject and scene appearance, motion, viewing positions, as well as action duration. 

We argue that most of the existing action recognition methods are not designed to handle such heterogeneity. Typically, these approaches are evaluated only on simple datasets involving a small number of action classes and videos recorded in lab-controlled environments~\cite{Irani05,Veeraraghavan06}. Furthermore, in the design of the action recognizer very little consideration is usually given to the computational cost which, as a result, is often very high. 

We believe that modern applications of action recognition demand scalable systems that can operate efficiently on large databases of unconstrained image sequences, such as YouTube videos. For this purpose, we identify three key-requirements to address: 1) the action recognition system must be able to handle the substantial variations of motion and appearance exhibited by realistic videos; 2) the training of each action classifier must have low-computational complexity and require little human intervention in order to be able to learn models for a large number of human actions; 3) the testing of the action classifier must be efficient so as to enable recognition in large repositories, such as video-sharing websites.

This work addresses these requirements by proposing a global video descriptor that yields state-of-the-art action recognition accuracy even with simple {\em linear} classification models. The feature entries of our descriptor are obtained by evaluating a set of movement classifiers over the video. Each of these classifiers is an exemplar-SVM~\cite{Malisiewicz11} trained on low-level features~\cite{Laptev:IJCV2005, WangEtAl:IJCV13} and optimized to separate a single positive video exemplar from an army of ``background'' negative videos. Because only one annotated video is needed to train an exemplar-SVM, our features can be learned with very little human supervision. The intuition behind our proposed descriptor is that it provides a semantically-rich description of a video by measuring the presence/absence of movements similar to those in the exemplars. Thus, a linear classifier trained on this representation will express a new action-class as a linear combination of the exemplar movements (which we abbreviate as EXMOVES). We demonstrate that these simple linear classification models produce surprisingly good results on challenging action datasets. In addition to yielding high-accuracy, these linear models are obviously very efficient to train and test, thus enabling {\em scalable} action recognition,  i.e., efficient recognition of many actions in large databases.

Our approach can be viewed as extending to videos the idea of classifier-based image descriptors~\cite{WangEtAl:ICCV2009, TorresaniEtAl:ECCV10, LiEtAl:NIPS2010, DengEtAl:CVPR2011} which describe a photo in terms of its relation to a set of predefined object classes. To represent videos, instead of using object classes, we adopt a set of movement exemplars. In the domain of action recognition, our approach is most closely related to the work of Sadanand and Corso~\cite{Sadanand12}, who have been the first to describe videos in terms of a set of actions, which they call the Action Bank. The individual features in Action Bank are computed by convolving the video with a set of predefined action templates. This representation achieves high accuracy on several benchmarks. However, the template-matching step to extract these mid-level features is very computationally expensive. As reported in~\cite{Sadanand12}, extracting mid-level features from a single video of UCF50~\cite{UCF101} takes a minimum of $0.4$ hours up to a maximum of $34$ hours. This computational bottleneck effectively limits the number of basis templates that can be used for the representation and constrains the applicability of the approach to small datasets. 

Our first contribution is to replace this prohibitively expensive procedure with a technique that is almost two orders of magnitude faster.  This makes our descriptor applicable to action recognition in large video databases, where the Action Bank framework is simply too costly to be used. The second advantage of our approach is that our mid-level representation can be built on top of any arbitrary spatial-temporal low-level features, such as appearance-based descriptors computed at interest points or over temporal trajectories. This allows us to leverage the recent advances in design of low-level features: for example, we show that when we use dense trajectories~\cite{WangEtAl:IJCV13} as low-level features, a simple linear classifier trained on the HMDB51 dataset using our mid-level representation yields a $41.6\%$ relative improvement in accuracy over the Action Bank built from the same set of video exemplars. Furthermore, we demonstrate that a linear classifier applied to our mid-level representation produces consistently much higher accuracy than the same linear model directly trained on the low-level features used by our descriptor.

Our EXMOVES are also related to Discriminative Patches~\cite{Jain13}, which are spatial-temporal volumes selected from a large collection of random video patches by optimizing a discriminative criterion. The selected patches are then used as a mid-level vocabulary for action recognition. Our approach differs from this prior work in several ways. As discussed in~\ref{efficienttesting}, each EXMOVE feature can be computed from simple summations over individual voxels. This model enables the use of  \emph{Integral Videos}~\cite{Yanke10}, which reduce dramatically the time needed to extract our features. Discriminative Patches cannot take advantage of the Integral Video speedup and thus they are much more computationally expensive to compute. This prevents their application in large-scale scenarios. On the other hand, Discriminative Patches offer the advantage that they are automatically mined, without any human intervention. EXMOVES require some amount of human supervision, although minimal (just one hand-selected volume per exemplar). In practice such annotations are inexpensive to obtain. In our experiments we show that EXMOVES learned from only 188 volumes greatly outperform Discriminative Patches using 4000 volumes.

\subsection{Related Work}
Many approaches to human action recognition have been proposed over the last decade. Most of these techniques differ in terms of the representation used to describe the video. An important family of methods is the class of action recognition systems using space-time interest points, such as Haris3D~\cite{Laptev:IJCV2005}, Cuboids~\cite{Piotr05}, and SIFT3D \cite{Scovanner07}. Efros~\emph{et al.} used optical flows to represent and classify actions~\cite{Efros03}. Klaser~\emph{et al.} extended HOG~\cite{Dalal06} to HOG3D by making use of the temporal dimension of videos~\cite{Klaser08}. Ke et al. learned volumetric features for action detection~\cite{Yanke10}. Wang and Suter proposed the use of silhouettes to describe human activities~\cite{Sutter07}. Recently, accurate action recognition has been demonstrated using dense trajectories and motion boundary descriptors~\cite{WangEtAl:IJCV13}.

On all these representations, a variety of classification models have been applied to recognize human actions: bag-of-word model~\cite{FeiFei07}, Metric Learning~\cite{Tran08}, Deep Learning~\cite{Quoc11}, Boosting-based approaches~\cite{Laptev08,Laptev07}.

Although many of these approaches have been shown to yield good accuracy on standard human action benchmarks, they are difficult to scale to recognition in large repositories as they involve complex 
feature representations or learning models, which are too costly to compute on vast datasets.  

\begin{figure*}[t]
\begin{center}
   \includegraphics[width=0.76\linewidth, natwidth=10.15in, natheight=6.72in]{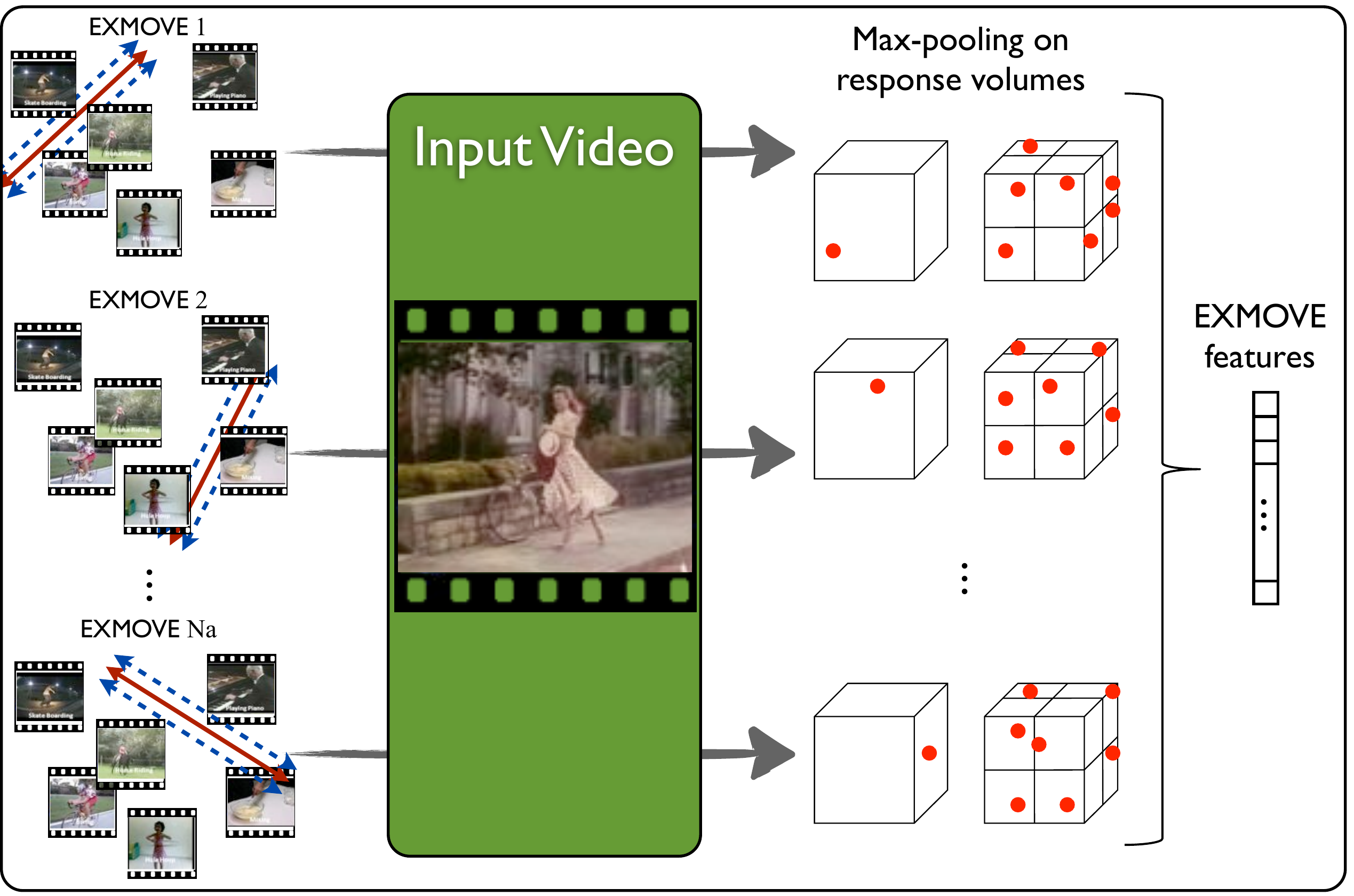}
\end{center}
\vspace{-0.4cm}
   \caption{{\bf Overview of our approach.} During an offline stage, a collection of exemplar-movement SVMs (EXMOVES) is learned. Each EXMOVE is trained using a single positive video exemplar and a large number of negative sequences. These classifiers are then used as mid-level feature extractors to produce a semantically-rich representation of videos. 
}
\label{fig:approach}
\end{figure*}

\section{Approach Overview}
We explain the approach at a high level using the schematic illustration in Figure~\ref{fig:approach}. During an offline stage, our method learns $N_a$ exemplar-movement SVMs (EXMOVES), shown on the left side of the figure. Each EXMOVE is a binary classifier optimized to recognize a specific action exemplar (e.g., an instance of ``biking'') and it uses histograms of quantized space-time low-level features for the classification. \LT{Note that in order to capture different forms of each activity, we use multiple exemplars per activity (e.g., multiple instances of ``biking''), each contributing a separate EXMOVE.} The set of learned EXMOVES are then used as mid-level feature extractors to produce an intermediate representation for any new input video: we evaluate each EXMOVE on subvolumes of the input video in order to compute the probability of the action at different space-time positions in the sequence. Specifically, we slide the subvolume of each EXMOVE exemplar at $N_s$ different scales over the input video. As discussed in section~\ref{efficienttesting}, this evaluation can be performed efficiently by using \emph{Integral Videos}~\cite{Yanke10}. Finally, for each EXMOVE, we perform max-pooling of the classifier scores within $N_p$ spatial-temporal pyramid volumes. Thus, for any input video this procedure produces a feature vector with $N_a \times N_s \times N_p$ dimensions. Because the EXMOVE features provide a semantically-rich representation of the video, even simple linear classification models trained on our descriptor achieve good action categorization accuracy.

\section{Exemplar-Movement SVMs (EXMOVES)}
\label{sec:midlevel}
Our EXMOVE classifiers are linear SVMs applied to histograms of quantized space-time low-level features calculated from subvolumes of the video. In section~\ref{sec:lowlevel} we describe the two space-time low-level descriptors used in our experiments, but any quantize-able appearance or motion features can be employed in our approach.

In principle, to train each SVM classifier we need a reasonable number of both positive and negative examples so as to produce good generalization. Unfortunately, we do not have many positive examples due to the high human cost of annotating videos. Thus, we resort to training each SVM using only one positive example, by extending to videos the exemplar-SVM model first introduced by Malisiewicz~\emph{et al.} for the case of still images~\cite{Malisiewicz11}. Specifically, for each positive exemplar, we manually specify a space-time volume enclosing the action of interest and excluding the irrelevant portions of the video. The histogram of quantized low-level space-time features contained in this volume becomes the representation used to describe the positive exemplar. Then, our objective is to learn a linear SVM that separates the positive exemplar from the histograms computed from all possible subvolumes of the same size in negative videos.

It may appear that training a movement classifier from a single example will lead to severe overfitting. However, as already noted in~\cite{Malisiewicz11}, exemplar-SVMs actually have good generalization as their decision boundary is tightly constrained by the millions of negative examples that the classifier must distinguish from the positive one. In a sense, the classifier is given access to an incredible amount of training examples to learn what the positive class is {\em not}. Furthermore, we use the exemplar-SVMs simply as mid-level feature extractors to find movements similar to the positive exemplar. Thus, their individual categorization accuracy is secondary. In other words, rather than applying the individual exemplar-SVMs as action recognizers, we use them collectively as building blocks to define our  action categorization model, in a role similar to the weak-learners of boosting techniques~\cite{ViolaJones:2001}.

\subsection{Low-level features used in EXMOVES} 
\label{sec:lowlevel}

Although any arbitrary low-level description of space-time points or trajectories can be used in our framework, here we experiment with the two following representations: \vspace{-5pt}
\begin{itemize}[noitemsep,nolistsep,leftmargin=*]
\item {\bf HOG-HOF-STIPs}. Given the input video, we first extract spatial-temporal interest points (STIPs)~\cite{Laptev:IJCV2005}. At each STIP we compute a Histogram of Oriented Gradients (HOG) and a Histogram of Flows (HOF)~\cite{Dalal06} using the implementation in~\cite{Laptev08}. We concatenate the HOG and the HOF descriptor to form a 162-dimensional vector representing the STIP. Finally, we run $k$-means on these vectors to learn a codebook of $D=5,000$ cluster centroids. Given the codebook, any space-time volume in a video is represented in terms of the histogram of codewords occurring within that  volume. We normalize the final histogram using the L$1$ norm. 
\item {\bf Dense Trajectories}. These are the low-level motion and appearance descriptors obtained from dense trajectories according to the algorithm described in~\cite{WangEtAl:IJCV13}. The trajectories are computed for non-stationary points using a median-filtered optical flow method and are truncated every 15 frames. Each trajectory is then described in terms of its shape (point coordinate features, $30$ dimensions), appearance (HOG features, $96$ dimensions), optical flow (HOF features, $108$ dimensions) and boundary motion (MBHx and MBHy features, $96$ dimensions each). As in~\cite{WangEtAl:IJCV13}, we learn a separate dictionary for each of these 5 descriptors. We use a codebook of $d=5,000$ cluster centroids for each descriptor. Thus, each space-time volume in a video is then represented as a vector of $D=25,000$ dimensions obtained by concatenating the $5$ histograms of trajectories occurring within that volume. We L$1$-normalize the final histogram.
\end{itemize}

\subsection{Learning EXMOVES}
The input for learning an EXMOVE consists of a positive video $\mathcal{V}^{+}$ containing a manually-annotated space-time $3$D box bounding the action of interest $\bx_E$, and thousands of negative videos $\mathcal{V}^{-}_{1..N}$ {\em without} action volume annotations. The only requirement on the negative videos is that they must represent action classes different from the category of the positive exemplar (e.g., if the exemplar contains the action {\em dancing}, we exclude dancing videos from the negative set). But this constraint can be simply enforced given action class labels for the videos, without the need to know the space-time volumes of these negative actions. \LT{For example, tagged Internet videos (e.g., YouTube sequences) could be used \DTR{to }as negative videos, by choosing action tags different from the activity of the positive exemplar.}

It is worth noting that different movement exemplars will have different $3$D box shapes. For example, we expect a walking action to require a tall volume while swimming may have a volume more horizontally elongated. As further discussed below, we maintain the original shape-ratio of the exemplar volume in both training and testing. This means that we look for only tall volumes when detecting walking, and short-and-wide volumes when searching for the swimming action.

Let $\bx_E$ be the manually-specified volume in the positive sequence $\mathcal{V}^{+}$.
Let us denote with $\bphi(\bx)$ the L1-normalized histogram of codewords (computed from either HOG-HOF-STIPs or Dense Trajectories) within a video volume $\bx$, i.e., $\bphi(\bx) = \frac{1}{c(\bx)}\left[c_1(\bx), \hdots, c_D(\bx)\right]^T$, where $c_i(\bx)$ is the number of codeword $i$ occurring in volume $\bx$, and $c(\bx)$ is the total number of codewords in $\bx$. Note that in the case of Dense Trajectories, each trajectory contributes 5 codewords into the histogram since it is quantized according to the $5$ separate dictionaries.

Adopting the exemplar-SVM method in~\cite{Malisiewicz11}, our exemplar-SVM training procedure learns a linear classifier $f(\bx)=\bw^T\bphi(\bx)+b$, by minimizing the following objective function:
\begin{multline}
\underset{\bw,b}{\mbox{min}} \ \|\bw\|^2 + C_1 \sum_{\bx\in\mathcal{V}^{+}~{\text s.t.} \frac{|\bx \cap \bx_E|}{|\bx_E|} \geq 0.5} h\bigl(\bw^T\phi(\bx)+b\bigr)\\
+ C_2 \sum_{i=1}^N \sum_{\bx\in\mathcal{V}^{-}_{i}}h\bigl(-\bw^T\bphi(\bx)-b\bigr)
\label{equ:exemplar_svm}
\end{multline}
where $h(s)=\max(0,1-s)$ is the hinge loss,  while $C_1$ and $C_2$ are pre-defined parameters that we set so as to equalize the unbalanced proportion of positive and negative examples. Note that the first summation in the objective involves subvolumes whose spatial overlap with $\bx_E$ is greater than $50\%$ and thus are expected to yield a positive score, while the second summation is over all negative subvolumes. Unfortunately, direct minimization of the objective in Eq.~\ref{equ:exemplar_svm} is not feasible since it requires optimizing the SVM parameters on a gigantic number of subvolumes. Thus, we resort to an alternation scheme similar to that used in~\cite{Malisiewicz11} and~\cite{FelzenszwalbEtAl:PAMI2010}: we iterate between 1) learning the parameters $(\bw,b)$ given an active set $S$ of negative volumes and 2) mining new negative volumes with the current SVM parameters. 

We first initialize the parameters of the classifier by traditional SVM training using the manually-selected volume $\bx_E$ as positive example and a randomly selected subvolumes from each of the other videos as negative example. At each iteration the current SVM is evaluated exhaustively on every negative video to find violating subvolumes, i.e., subvolumes yielding an SVM score below exceeding $-1$. These subvolumes are added as negative examples to the active set $S$ to be used in the successive iterations of SVM learning. Furthermore, our training procedure adds as positive examples the subvolumes of $\mathcal{V}^{+}$ that have spatial overlap with $\bx_E$ greater than $50\%$ and SVM score below $1$. We stop the iterative alternation between these two steps when either no new subvolumes are added to the active set or a maximum number of iterations $M$ is reached. In our implementation we use $M=10$, but we find that in more than 85\% of the cases, the learning procedure converges before reaching  this maximum number of iterations.

The pseudocode of our learning procedure is given in Algorithm~1. Lines $1-3$ initialize the active set. The function ${\tt svm\_training}$ in line 5 learns a traditional binary linear SVM using the labelled examples in the active set. Note that we found that at each iteration we typically have millions of subvolumes violating the constraints (lines 7-11). In order to maintain the learning of the SVM feasible, in practice we add to the active set only the volumes that yield the largest violations in each video, for a maximum of $k^{-}=3$ per negative video and $k^{+}=10$ for the positive video.

\renewcommand{\algorithmicrequire}{\textbf{Input:}}
\renewcommand{\algorithmicensure}{\textbf{Output:}}
\begin{algorithm}[t]		
{\small
\caption{EXMOVE training}
\begin{algorithmic}[1]
\label{algo:alg1}    
    \REQUIRE A set of negative videos  $\{\mathcal{V}^-_1, \hdots, \mathcal{V}^-_N\}$ and a manually-selected volume $\bx_E$ in exemplar video $\mathcal{V}^+$.
    \ENSURE Parameters $(\bw, b)$ of exemplar-SVM.
    \STATE $S \leftarrow \{(\bx_E, +1)\}$ 
    \FOR  {$i = 1$ to $N$}  
    \STATE $S \leftarrow S \cup \{(\bx_i, -1)\}$ with $\bx_i$ randomly chosen from $\mathcal{V}^-_i$        
    \ENDFOR   
    \FOR  {$iter = 1$ to $M$}  
    	\STATE $(\bw, b) \leftarrow{\tt svm\_training}(S)$
	\STATE $S_{old} \leftarrow S$
	\FORALL {$\bx$ in $\mathcal{V}^+$ s.t. $\bw^T \bx + b < 1$ \& $\frac{|\bx \cap \bx_E|}{|\bx_E|} > 0.5$} 
		\STATE $S \leftarrow S \cup \{(\bx, +1)\}$ \textsl{//false negative}
       	\ENDFOR
    	\FOR  {$i = 1$ to $N$} 
        		\FORALL {$\bx$ in $\mathcal{V}^-_i$ s.t. $\bw^T \bx + b > -1$ } 
		\STATE $S \leftarrow S \cup \{(\bx, -1)\}$ \textsl{//false positive}
        		\ENDFOR
	\ENDFOR
	\IF {$S_{old} = S$}
		 \STATE break
	\ENDIF
    \ENDFOR    
\end{algorithmic}
}
\end{algorithm}

\subsection{Calibrating the ensemble of EXMOVES} The learning procedure described above is applied to each positive exemplar independently to produce a collection of EXMOVES. However, because the exemplar classifiers are trained disjointly, their score ranges and distributions may vary considerably. A standard solution to this problem is to calibrate the outputs by learning for each classifier a function that converts the raw SVM score into a proper posterior probability compatible across different classes. To achieve this goal we use the procedure proposed by Platt in~\cite{Platt:ALMC99}: for each exemplar-SVM $(\bw_E,b_E)$ we learn parameters $(\alpha_E, \beta_E)$ to produce calibrated probabilities through the sigmoid function $g(\bx; \bw_E, b_E, \alpha_E, \beta_E) = 1/[1 + \exp(\alpha_E (\bw_E^T \bx + b_E) + \beta_E)]$. The fitting of parameters $(\alpha_E, \beta_E)$ is performed according to the iterative optimization described in~\cite{Platt:ALMC99} using as labeled examples the positive/negative volumes that are in the active set at the completion of the EXMOVE training procedure. As already noted in~\cite{Malisiewicz11}, we also found that this calibration procedure yields a significant improvement in accuracy since it makes the range of scores more homogeneous and diminishes the effect of outlier values. 

\subsection{Efficient computation of EXMOVE scores}
\label{efficienttesting}
Although replacing the template matching procedure of Action Bank with linear SVMs applied to histograms of space-time features yields a good computational saving, this by itself is still not fast enough to be used in large-scale datasets due to the exhaustive sliding volume scheme. In fact, the sliding volume scheme is used in both training and testing. In training, we need to slide the current SVM over negative videos to find volumes violating the classification constraint. In testing, we need to slide the entire set of EXMOVE classifiers over the input video in order to extract the mid-level features for the subsequent recognition. Below, we describe a solution to speed up the sliding volume evaluation of the SVMs.

Let $\mathcal{V}$ be an input video of size $R \times C \times T$. Given an EXMOVE with parameters $(\bw_E, b_E)$, we need to efficiently evaluate it over all subvolumes of $\mathcal{V}$ having size equal to the positive exemplar subvolume $\bx_E$ (in practice, we slide the subvolume at $N_s$ different scales but for simplicity we illustrate the procedure assuming we use the original scale). It is worth noting that the branch-and-bound method of Lampert \emph{et al.}~\cite{Lampert09} cannot be applied to our problem because it can only find the subwindow maximizing the classification score while we need the scores of all subvolumes; moreover it requires  unnormalized histograms. 

Instead, we use integral videos~\cite{Yanke10} to efficiently compute the  EXMOVE score for each subvolume. An integral video is a volumetric data-structure having size equal to the input sequence (in this case $R \times C \times T$). It is useful to speed up the computation of functions defined over subvolumes and expressed as cumulative sums over voxels, i.e, functions of the form $H(\bx) = \sum_{(r,c,t) \in  \bx} h(r,c,t)$, where $(r,c,t)$ denotes a space-time point in volume $\bx$ and $h$ is a function over individual space-time voxels. The integral video for $h$ at point $(r,c,t)$ is simply an accumulation buffer $B$ storing the sum of $h$ over all voxels at locations less than or equal to $(r,c,t)$, i.e., $B(r,c,t) = \sum_{r' \leq r} \sum_{c' \leq c} \sum_{t' \leq t} h(r',c',t')$. This buffer can be built with complexity linear in the video size. Once built, it can be used to compute $H(\bx)$ for any subvolume $\bx$ via a handful of additions/subtractions of the values in $B$. 

In our case, the use of integral video is enabled by the fact that the classifier score can be expressed in terms of cumulative sums of individual point contributions, as we illustrate next. For simplicity we describe the procedure assuming that $\bphi(\bx)$ consists of a single histogram (as is the case for HOG-HOF-STIPs) but the method is straightforward to adapt for the scenario where $\bphi(\bx)$ is the concatenation of multiple histograms (e.g., the 5 histograms of Dense Trajectories). Let us indicate with $P(\bx)$ the set of quantized low-level features (either STIPs or Dense Trajectories) included in subvolume $\bx$ of video $\mathcal{V}$ and let $i_p$ be the codeword index of a point $p \in P(\bx)$. Then we can rewrite the classification score of exemplar-SVM $(\bw, b)$ on a subvolume $\bx$ as follows (we omit the constant bias term $b$ for brevity):
\begin{eqnarray}
\bw^T \bphi(\bx) = \frac{1}{c(\bx)}\sum_{i=1}^D w_i c_i(\bx) = \frac{\sum_{p \in P(\bx)} w_{i_p}}{\sum_{p \in P(\bx)}1}~.
\label{equ:evaluation2}
\end{eqnarray}
Equation~\ref{equ:evaluation2} shows that the classifier score is expressed as a ratio where both the numerator and the denominator are computed as sums over individual voxels. Thus, the classifier score for any $\bx$ can be efficiently calculated using two integral videos (one for the numerator, one for the denominator), without ever explicitly computing the histogram $\bphi(\bx)$ or the inner product between $\bw$ and $\bphi(\bx)$. In the case where $\bphi(\bx)$ contains the concatenation of multiple histograms, then we would need an integral video for each of the histograms (thus 5 for Dense Trajectories), in addition to the common integral video for the denominator.
 
 \begin{table*}[th!]
\begin{center}
{\footnotesize
\begin{tabular}{|c|c|c|c|c|c|c|}
\hline
 & & & \multicolumn{4}{c}{\bf Datasets}\\
\shortstack{{\bf Low-level}\\ {\bf features}} & \shortstack{{\bf Mid-level}\\{\bf descriptor}}  & \LT{\shortstack{{\bf Descriptor}\\{\bf dimensionality}}} & HMDB51 & Hollywood-2 & UCF50 & UCF101(part 2) \\
 \hline \hline
3D Gaussians & Action Bank & 44,895 & 26.9 & n/a & 57.9 & n/a \\
\hline \hline
HOG3D & Discriminative Patches & \DT{9,360} & n/a & n/a & 61.2 & n/a \\
\hline
\hline
\multirow{2}{*}{HOG-HOF-STIPs} & BOW & \DT{5,000} & 20.0 & 32.6 & 52.8 & 49.1 \\
& EXMOVES & 41,172 & 27.7 & 44.7 & 63.4 & 57.2 \\
\hline \hline
\multirow{2}{*}{Dense Trajectories} & BOW & 25,000 & 34.4 & 43.7 & 81.8 & 60.9\\
& EXMOVES & 41,172 & {\bf 41.9} & {\bf 56.6} & {\bf 82.8} & {\bf 71.6} \\
\hline
\end{tabular}
}
\vspace{-0.1cm}
\caption{Comparison of recognition accuracies on four datasets. The classification model is an efficient linear SVM applied to 4 distinct global mid-level descriptors: Action Bank~\cite{Sadanand12}, Discriminative Patches~\cite{Jain13}, Histogram of Space-Time Visual Words (BOW) and our EXMOVES. We consider two different low-level features to build BOW and EXMOVES: HOG-HOF-STIPs and Dense Trajectories. Our EXMOVES achieve the best recognition accuracy on all four datasets using Dense Trajectories, and greatly outperform the BOW descriptor for both our choices of low-level features, HOG-HOF-STIPs and Dense Trajectories.\vspace{-0.3cm}}
\label{tab:allrec}
\end{center}
\end{table*}

\DTR{\LT{Du, can you please double-check the dimensionalities of all descriptors in Table 1?}}\DTR{Checked}

\section{Experiments}

\subsection{Experimental setup}\vspace{-5pt}
\myparagraph{Implementation details of EXMOVE training.}Since our approach shares many similarities with Action Bank, we adopt training and design settings similar to those used in~\cite{Sadanand12} so as to facilitate the comparison between these two methods. Specifically, our EXMOVES are learned from the same set of UCF50~\cite{UCF101} videos used to build the Action Bank templates. This set consists of $188$ sequences spanning a total of $50$ actions. Since the Action Bank volume annotations are not publicly available, we manually selected the action volume $\bx_E$ on each of these exemplar sequences to obtain $N_a = 188$ exemplars. As negative set of videos we use the remaining $6492$ sequences in the UCF50 dataset: for these videos no manual labeling of the action volume is available nor it is needed by our method. Action Bank also includes 6 templates taken from other sources but these videos have not been made publicly available; it also uses 10 templates taken from the KTH dataset. However, as the KTH videos are lower-resolution and contain much simpler actions compared to those in UCF50, we have not used them to build our EXMOVES. In the experiments we show that, while our descriptor is defined by a smaller number of movement classifiers ($188$ instead of $205$), the recognition performance obtained with our mid-level features is consistently on par with or better than Action Bank.

\myparagraph{Parameters of EXMOVE features.}In order to compute the EXMOVE features from a new video, we perform max-pooling of the EXMOVE scores using a space-time pyramid based on the same settings as those of Action Bank, i.e., $N_s=3$ scaled versions of the exemplar volume $\bx_E$ (the scales are 1, 0.75, 0.5), and $N_p=73$ space-time volumes obtained by recursive octree subdivision of the entire video using 3 levels (this yields 1 volume at level 1, 8  subvolumes at level 2, 64 subvolumes at level 3). Thus, the final dimensionality of our EXMOVE descriptor is $N_a \times N_s \times N_p = 41,172$.

\myparagraph{Action classification model.}All our action recognition experiments are performed by training a one-vs-the-rest linear SVM on the EXMOVES extracted from a set of training videos. We opted for this classifier as it is very efficient to train and test, and thus it is a suitable choice for the scenario of large-scale action recognition that we are interested in addressing. The hyperparameter $C$ of the SVM is tuned via cross-validation for all baselines, Action Bank, and our EXMOVES.

\myparagraph{Test datasets.}We test our approach on the following large-scale action recognition datasets:
\begin{enumerate}[noitemsep,nolistsep,leftmargin=*]
\item HMDB51~\cite{Kuehne11}: It consists of $6849$ image sequences collected from movies as well as YouTube and Google videos. They represent 51 action categories. The results for this dataset are presented using $3$-fold cross validation on the $3$ publicly available training/testing splits.
\item Hollywood-2~\cite{Marszalek09}: This dataset includes over 20 hours of video, subdivided in $3669$ sequences, spanning 12 action classes. We use the publicly available split of training and testing examples.
\item UCF50: This dataset contains 6676 videos taken from YouTube for a total of 50 action categories. This dataset was used in~\cite{Sadanand12} and~\cite{Jain13} to train and evaluate Action Bank and Discriminative Patches. 
\item UCF101~\cite{UCF101} (part 2): UCF101 is a superset of UCF50. For this test we only use videos from action classes 51 to 101 (from now on denoted as part 2), thus omitting the above-mentioned classes and videos of UCF50. This leaves a total of 6851 videos and 51 action classes. We report the accuracy of 25-fold cross validation using the publicly available training/testing splits.

\end{enumerate}

\subsection{Action recognition}

\myparagraph{Comparison of recognition accuracies.}~We now present the classification performance obtained with our features on the four benchmarks described above. We consider in our comparison three other mid-level video descriptors that can be used for action recognition with linear SVMs: Action Bank~\cite{Sadanand12}, Discriminative Patches~\cite{Jain13} as well as histograms of visual words (BOW) built for the two types of low-level features that we use in EXMOVES, i.e., HOG-HOF-STIPs and Dense Trajectories. \LT{As in~\cite{WangEtAl:IJCV13}, we use a dictionary of 25,000 visual words for Dense Trajectories \DT{and 5,000 visual words for HOG-HOF-STIPs}}. Due to the high computational complexity of the extraction of Action Bank features, however, we were unable to test this descriptor on the large-scale datasets of Hollywood-2 and UCF101. For Discriminative Patches, we can only report accuracy on UCF50 since this is the only large-scale dataset on which they were tested in~\cite{Jain13} and no software to compute these features is available. 

The accuracies achieved by the different descriptors are summarized in Table~\ref{tab:allrec}. From these results we see that our EXMOVE descriptor built from Dense Trajectories yields consistently the best results across all four datasets. Furthermore, EXMOVES gives always higher accuracy than BOW built from the same low-level features, for both HOG-HOF-STIPs and Dense Trajectories. The gap is particularly large on challenging datasets such as Hollywood-2 and HMDB51. This underscores the advantageous effect of the movement exemplars to which we compare the input video in order to produce the EXMOVE features. 

Table~\ref{tab:ucf13} lists the individual action recognition accuracies for the same subset of 13 classes analyzed in~\cite{Jain13}. We see that EXMOVES give the highest accuracy on 10 out of these 13 action categories.

\myparagraph{Computational cost of mid-level feature extraction.}We want to emphasize that although our EXMOVES are based on a subset of the exemplars used to build Action Bank, they always generate equal or  higher accuracy. Furthermore, our approach does so with a speedup of almost two-orders of magnitude in feature extraction: Table~\ref{tab:runtimes} reports the statistics of the runtime needed to extract EXMOVES  and Action Bank. We used the software provided by the authors of~\cite{Sadanand12} to extract Action Bank features from input videos. Due to large cost of Action Bank extraction, we collected our runtime statistics on the smaller-scale 
UT-I~\cite{UT-Interaction-Data} dataset, involving only 120 videos. Runtimes were measured on a single-core Linux machine with a CPU @ 2.66GHz. The table reports the complete time from the input of the video to the output of the descriptor, inclusive of the time needed to compute low-level features. The extraction of EXMOVES is on average over $70$ times faster than for Action Bank when using HOG-HOF-STIPs and $11$ times faster when using Dense Trajectories. We can process the entire UT-Interaction dataset with HOG-HOF-STIPs using a single CPU in $14$ hours; extracting the Action Bank features on the same dataset would take $41$ days.

We were unable to collect runtime statistics for Discriminative Patches due to the unavailability of the software. However, we want to point out that this descriptor uses many more patches than EXMOVES (\DT{1040} \DTR{4000} instead of 188) and it cannot use the Integral Video speedup.

\begin{table}
{\footnotesize
\begin{tabular}{|l|c|c|c|}
\hline 
Action Class & \shortstack{Action\\Bank} & \shortstack{Discriminative\\Patches} & EXMOVES \\ 
\hline 
\hline 
Basketball & 53.84 & 50.00 & {\bf 56.93} \\
Clean and Jerk & 85.00 & {\bf 95.65} & 91.07\\
Diving & 78.79 & 61.29 & {\bf 96.08} \\
Golf Swing & {\bf 90.32} & 75.86 & 90.14 \\
High Jump & 38.46 & 55.56 & {\bf 81.30} \\
Javeline Throw & 45.83 & 50.00 & {\bf 73.50} \\
Mixing & 42.85 & 55.56 & {\bf 97.16} \\
PoleVault & 60.60 & 84.37 & {\bf 94.38} \\
Pull Up & 91.67 & 75.00 & {\bf 96.00}\\
Push Ups & 85.00 & 86.36 & {\bf 91.18} \\
Tennis Swing & 44.12 & 48.48 & {\bf 85.03} \\
Throw Discus & 75.00 & 87.10 & {\bf 93.13} \\
Volleyball Spiking & 43.48 & {\bf 90.90} & 89.66\\
\hline
\hline 
Mean Classification & 64.23 & 70.47 & {\bf 87.35}\\
\hline 
\end{tabular}\vspace{-2mm}
}
\caption{Recognition accuracies of our EXMOVES (applied to Dense Trajectories) compared with
those of Action Bank and Discriminative Patches using the same subset of 13 action classes from UCF50 considered in~\cite{Jain13}.}
\label{tab:ucf13}
\end{table}

\begin{table}[t]

{\footnotesize
\begin{center}
\begin{tabular}{|c||c|c|c||c|}

\hline
Descriptor & \multicolumn{3}{|c||}{\shortstack{Extraction time\\per video (minutes)}} & \LT{\shortstack{\# frames\\per second}}\\ 
 & mean & max & min & mean\\
\hline \hline
Action Bank & 495 & 1199 & 132 & 0.012\\
\hline
\shortstack{\\EXMOVES\\w/ HOG-HOF-STIPs} & 7 & 16 & 3 & 0.82\\
\hline
\shortstack{\\EXMOVES\\w/ Dense Trajectory} & 43 & 70 & 29 & 0.13\\
\hline
\end{tabular}
\end{center}\vspace{-0.1cm}
\caption{Statistics of time needed to extract the mid-level descriptors Action Bank and EXMOVES. The time needed to extract EXMOVES features for the entire UT-I dataset using a single CPU is only 14 hours; instead,  it would take more than 41 days to compute Action Bank descriptors for this dataset.\vspace{-0.1cm}} 
\label{tab:runtimes}
}
\end{table}

%
%

\myparagraph{Computational cost of action recognition.}Finally, we would like to point out that as shown in Table~\ref{tab:allrec}, the accuracies achieved  by an {\em efficient linear} SVM trained on EXMOVES are very close to the best published results of~\cite{WangEtAl:IJCV13}, which instead were obtained with a much more computationally expensive model, not suitable for scalable action recognition: they report a top-performance of $46.6\%$ and $58.2\%$ on HMDB51 and Hollywood-2, respectively, using an expensive non-linear SVM with an RBF-$\chi^2$ kernel applied to BOW of Dense Trajectories. In our experiments we found that training a linear SVM on EXMOVES for one of the HMDB51 classes takes only 6.2 seconds but learning a kernel-SVM on BOW of Dense Trajectories requires 25 minutes (thus overhead is 250X); the testing of our linear SVM on a video takes only 7 milliseconds, while the nonlinear SVM is on average more than two orders of magnitude slower. Its cost depends on the on the number of support vectors, which varies from a few hundreds to several thousands. Nonlinear SVMs also need more memory to store the support vectors.

\myparagraph{Varying the number of exemplars.}In this experiment we study how the accuracy of our method changes as a function of the number of EXMOVES used in the descriptor. Starting from our complete feature vector defined by $N_a=188$ exemplars and having dimensionality $N_a \times N_s \times N_p = 41,172$, we recursively apply a feature selection procedure that eliminates at each iteration one of the EXMOVE exemplars and removes its associated $N_s \times N_p$ features from the descriptor. We apply a variant of multi-class Recursive Feature Elimination~\cite{ChapelleKeerthi:ASA2008} to determine the EXMOVE to eliminate at each iteration. This procedure operates as follows: given a labeled training set of video examples for $K$ classes, at each iteration we retrain the one-vs-the-rest linear SVMs for all $K$ classes using the current version of our feature vector and then we remove from the descriptor the EXMOVE that is overall ``least used'' by the $K$ linear classifiers by looking at the average magnitude of the SVM parameter vector $\bw$ for the different EXMOVE sub-blocks.

\begin{figure}
{\includegraphics[width=0.9\linewidth, natwidth=7.25in, natheight=5.75in]{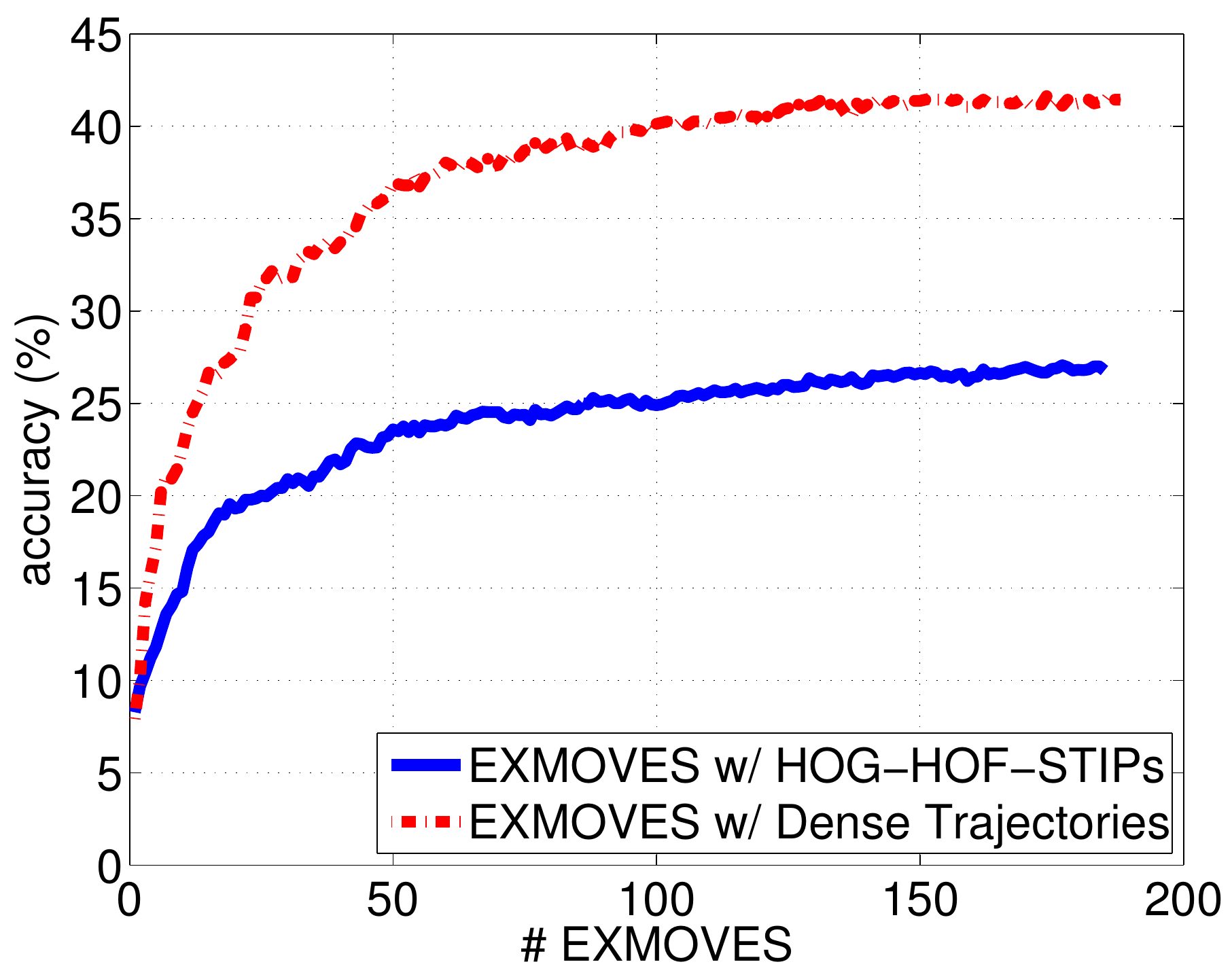}}\vspace{-2mm}
{
\caption{Accuracy on HMDB51 as a function of the number of EXMOVES. We use Recursive Feature Elimination to reduce the number of  EXMOVES. The accuracy remains near the state-of-the-art even when using only $100$ exemplars.}\label{fig:accvsnumEXMOVES}
}
\vspace{-3mm}
\end{figure}

We perform this analysis on the HDMB51 dataset using both HOG-HOF-STIPs and Dense Trajectories as low-level features for EXMOVES.  Figure~\ref{fig:accvsnumEXMOVES} reports the 3-fold cross-validation error as a function of the number of EXMOVES used in our descriptor. Interestingly, we see that the accuracy remains close to the top-performance even when we reduce the number of exemplars to only 100. This suggests a certain redundancy in the set of movement exemplars. The accuracy begins to drop much more rapidly when fewer than 50 exemplars are used. 
\LTR{In our supplementary material we list the first 20 EXMOVES removed by the feature elimination procedure (i.e., the least useful features) as well as the last 20 EXMOVES to be discarded (i.e., the most useful features).}

\section{Conclusions}
We have presented an approach for efficient large-scale human action recognition. It centers around the learning of a mid-level video representation that enables state-of-the-art accuracy with efficient linear classification models. Experiments on large-scale action recognition benchmarks show the accuracy and efficiency of our approach.

Our mid-level features are produced by evaluating a predefined set of movement classifiers over the input video. An important question we plan to address in future work is: how many mid-level classifiers do we need to train before accuracy levels off? Also, what kind of movement classes are particularly useful as mid-level features? Currently, we are restricted in the ability to answer these questions by the scarceness of labeled data available, in terms of both number of video examples but also number of action classes. An exciting avenue to resolve these issues is the design of methods that can learn robust mid-level classifiers from weakly-labeled data, such as YouTube videos. 

\LTR{The software implementing our approach and all the data used in the experiments will be released upon
paper acceptance.}
\LT{Additional material including software to extract EXMOVES from videos is available at~{\small\url{http://vlg.cs.dartmouth.edu/exmoves}}.}

\LT{
\section*{Acknowledgments}
Thanks to Alessandro Bergamo for assistance with the experiments. This research was funded in part by NSF CAREER award IIS-0952943 and NSF award CNS-1205521.}

{\small
\bibliographystyle{icml2014}
\bibliography{refs}
}

\end{document}